\pdfoutput=1

\documentclass[11pt]{article}

\usepackage[preprint]{acl}

\usepackage{times}
\usepackage{latexsym}

\usepackage[T1]{fontenc}

\usepackage[utf8]{inputenc}

\usepackage{microtype}

\usepackage{inconsolata}

\usepackage{graphicx}
\usepackage{amssymb}
\usepackage{booktabs}
\usepackage{amsmath}

\title{Structured Information Matters:\\ Explainable ICD Coding with Patient-Level Knowledge Graphs}

\author{
  Mingyang Li\textsuperscript{1}, 
  Viktor Schlegel\textsuperscript{2,1}, 
  Tingting Mu\textsuperscript{1}, 
  Warren Del-Pinto\textsuperscript{1}, 
  Goran Nenadic\textsuperscript{1} \\
  \textsuperscript{1}The University of Manchester, 
  \textsuperscript{2}Imperial College London \\
  \texttt{\{mingyang.li, Tingting.Mu, warren.del-pinto, gnenadic\}@manchester.ac.uk}, \\
  \texttt{v.schlegel@imperial.ac.uk}
}

\begin{document}
\maketitle
\begin{abstract}
Mapping clinical documents to standardised clinical vocabularies 
is an important task, as it provides structured data for information retrieval and analysis, which is essential to clinical research, hospital administration and improving patient care. 
However, manual coding is both difficult and time-consuming, making it impractical at scale. Automated coding can potentially alleviate this burden, improving the availability and accuracy of structured clinical data. The task is difficult to automate, as it requires mapping to high-dimensional and long-tailed target spaces, such as the International Classification of Diseases (ICD). 
While external knowledge sources have been readily utilised to enhance output code representation, the use of external resources for representing the input documents has been underexplored. In this work, we compute a structured representation of the input documents, making use of document-level knowledge graphs (KGs) that provide a comprehensive structured view of a patient's condition. The resulting knowledge graph efficiently represents the patient-centred input documents with 23\% of the original text while retaining 90\% of the information. We assess the effectiveness of this graph for automated ICD-9 coding by integrating it into the state-of-the-art ICD coding architecture PLM-ICD. Our experiments yield improved Macro-F1 scores by up to 3.20\% on popular benchmarks, while improving training efficiency. We attribute this improvement to different types of entities and relationships in the KG, and demonstrate the improved explainability potential of the approach over the text-only baseline. 
\end{abstract}

\section{Introduction}

\begin{figure}[t]
\centering
\includegraphics[width=0.95\columnwidth]{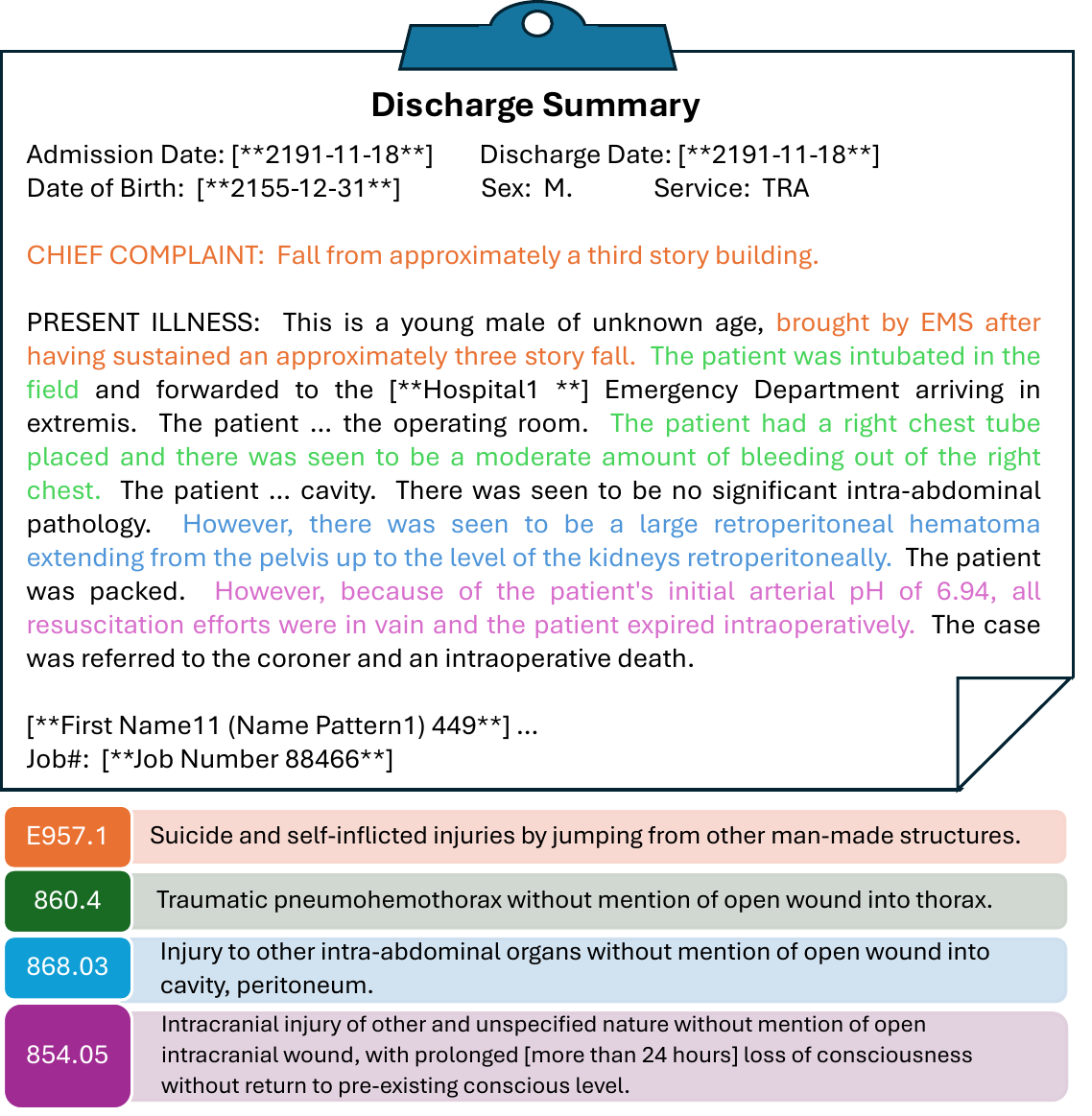} 
\caption{Example of ICD Coding over MIMIC-III. The discharge summary (HADM ID: 104128) is annotated with four ICD codes.}
\label{fig1}
\end{figure}

Clinical coding is the process of allocating standardized codes to diagnoses, treatments, procedures, and medical services detailed in patient electronic records or paper notes. This multi-label classification task offers advantages across various domains, including audit procedures, decision support systems and medical billing \cite{blundell2023health}. Various coding systems are designed to encode specific information within patient records. Our work focuses on the International Classification of Diseases (ICD-9) \cite{world1978international}, a widely recognized coding system that holds a pivotal role in encoding diagnostic and procedural information. This process is commonly known as ICD coding. An example is shown in Figure \ref{fig1}.

Manual code assignment is typically costly, labor-intensive, and error-prone \cite{nguyen2018computer}. 
In recent years, automated clinical coding, powered by cutting-edge deep learning techniques, has significantly advanced the field, improving accuracy, increasing efficiency, and reducing overall costs \cite{ji2022unified, teng2022review}.

The main challenge in clinical coding arises from the extremely imbalanced distribution of the label space. For instance, in the case of MIMIC-III \cite{johnson2016mimic}, there are 8,692 unique ICD-9 codes, of which 4,115 codes (47.3\%) occur fewer than 6 times \cite{yang2022knowledge}. 
Considering this long-tailed distribution of codes, previous work has explored integrating diverse external knowledge to enhance the representation of codes and patients. Among these external knowledge sources, knowledge graphs play an important role in improving the performance of ICD coding by providing not only semantic information but also structured information. However, most research focuses on representing ICD codes through various graphs that are built based on these codes themselves \cite{rios2018few, xie2019ehr, cao2020hypercore, lu2020multi, song2021generalized, michalopoulos2022icdbigbird}. Efforts to construct patient-level knowledge graphs remain largely underexplored in both ICD coding and the broader clinical domain.

The patient-level knowledge graph offers an intuitive representation and visualization of a patient’s clinical condition, providing healthcare professionals with valuable insights. Meaningful causal relationships between entities, such as symptoms that support a diagnosis, tests performed, and treatments derived from these findings, enable patient-level knowledge graphs to facilitate more efficient decision-making for physicians and medical staff. However, critical questions remain unanswered: what elements should constitute a patient’s knowledge graph, including problems, symptoms, tests, treatments, drugs, dosages, and frequencies? And how to evaluate the quality of such graphs and assess their utility and impact on tasks such as patient-level classification and explainability?

To the best of our knowledge, \citet{yuan2021graph} is the only work which proposes a medical graph specifically designed for individual patients in ICD coding task. The graph integrates a disease hierarchy based on ICD-10 and a causal graph of diseases. Entities in the causal graph, including symptoms, signs, and diseases are identified from documents using the named-entity recognition (NER) technique. The model also leverages GCN to represent the nodes in the graph. It enhances the patient representation by integrating it with the raw clinical text and patient information. However, it does not cover a wide range of entity categories and capture the diverse relationships among them, which can provide a more comprehensive understanding about a patient's medical history. Additionally, this work lacks a systematic evaluation of graph quality and an analysis of the determination of its constituent components.

To close these gaps, we construct patient-level knowledge graphs that provide a wide range of entity types and relationships. This comprehensive graph offers explicit context to a patient's situation, by providing diagnostic, posology, anatomical and the temporal information of clinical events identified in the patient records. 
We integrate this patient-level knowledge graph into the state-of-the-art ICD coding architecture, PLM-ICD~\cite{huang2022plm}, demonstrating improved coding performance.

The contributions of this work are:

    \emph{(i)} We develop a comprehensive patient-level knowledge graph encompassing a wide coverage of 14 distinct entity types connected by five types of relationships. We evaluate the informativeness of the graph by measuring the information loss relative to the patient notes from which the graph is retrieved. Our results demonstrate that the knowledge graph effectively distills essential information from patient notes into a more concise and structured format, achieving a significant reduction in size—extracting only 23\% of the original content—while retaining 90\% of the information. This represents a \textbf{\textit{Statistical Perspective}} for evaluating the quality of the graph.
    
    \emph{(ii)} We conducted experiments to assess the effectiveness of integrating graph representations into ICD coding. 
    The results demonstrate that the additional structured information provided by the graph enhances coding performance, improving the F1-score by 1.36\% compared to the base model. This improvement is significant for this developed task. This also serves as an evaluation of the patient-level knowledge graph from a \textbf{\textit{Representational Perspective}}, capturing both semantic and structural information.
    
    \emph{(iii)} We address the question of `\textbf{\textit{What elements should constitute a patient’s knowledge graph?}}' through an ablation study from two evaluation perspectives. We analyse the impact of various types of entities and relationships on the information retaining and coding performance.
    
    \emph{(iv)} We perform a case study and showcase the model's ability to offer high-quality explanations by providing accurate and concise evidence which supports the model's prediction.




\section{Related Work}

\paragraph{Architecture}
Over the past decade, the field of clinical coding has witnessed significant advancements, evolving from traditional rule-based methods \cite{pereira2006construction, crammer2007automatic} to advanced machine learning and deep learning approaches. Researchers have recently explored the application of cutting-edge NLP techniques, including attention mechanisms and transformer models. The architecture of these models has become increasingly sophisticated, with common architecture incorporating CNN-based \cite{mullenbach2018explainable}, LSTM-based \cite{catling2018towards}, and transformer-based encoders \cite{zhang2020bert, chalkidis2020empirical, ji2021does}, often paired with label-wise attention layers \cite{vu2020label, sun2021multitask, dong2021explainable, liu2021effective, van2022patient}. Recent studies also highlight the challenge of efficiently applying transformer models to represent the inherently lengthy clinical documents. These approaches leverage transformers handling long sequences, notably Longformer \cite{yang2022knowledge} and BigBird \cite{michalopoulos2022icdbigbird}.
\paragraph{External Knowledge Representations}
A major challenge in this field is classifying within a large target space, where the distribution of codes is highly uneven, commonly described as a `big-head long-tail' distribution. This imbalance hinders the model's effectiveness in recognising patterns associated with categories with few samples. To address this issue, researchers have turned to external knowledge to enhance the representations of both patients and codes. For patient representation, this includes data augmentation \cite{falis2022horses, song2021generalized} and knowledge graphs \cite{yuan2021graph}. In terms of code representation, external knowledge is drawn from code descriptions \cite{feucht2021description}, synonyms \cite{yuan2022code}, relevant documents \cite{wang2022novel}, code hierarchy \cite{falis2019ontological, yang2022knowledge}, synthetic data \cite{falis2022horses}, and knowledge graphs.
\paragraph{Knowledge Graph in ICD Coding}
\citet{rios2018few} represents ICD codes using their hierarchical structure, applying two layers of graph convolutional networks (GCN) to leverage this structured knowledge.  \citet{song2021generalized} improves this model by replacing the GCN with graph gated recurrent neural networks (GRNN) \cite{li2015gated}.
\citet{cao2020hypercore} introduces Co-Graph, which models co-occurrence correlations between codes. 
This graph is represented by its adjacency matrix and GCN. \citet{lu2020multi} constructs three types of graphs: a label hierarchy graph of class taxonomy, a semantic similarity graph derived from code descriptions, and a code co-occurrence graph similar to the approach in \citet{cao2020hypercore}. \citet{michalopoulos2022icdbigbird} establishes connections between codes using normalized point-wise mutual information and also employs GCN to capture the representations of codes from this graph.

\section{Methodology}

\begin{figure*}[t]
\centering
\includegraphics[width=0.96\textwidth]{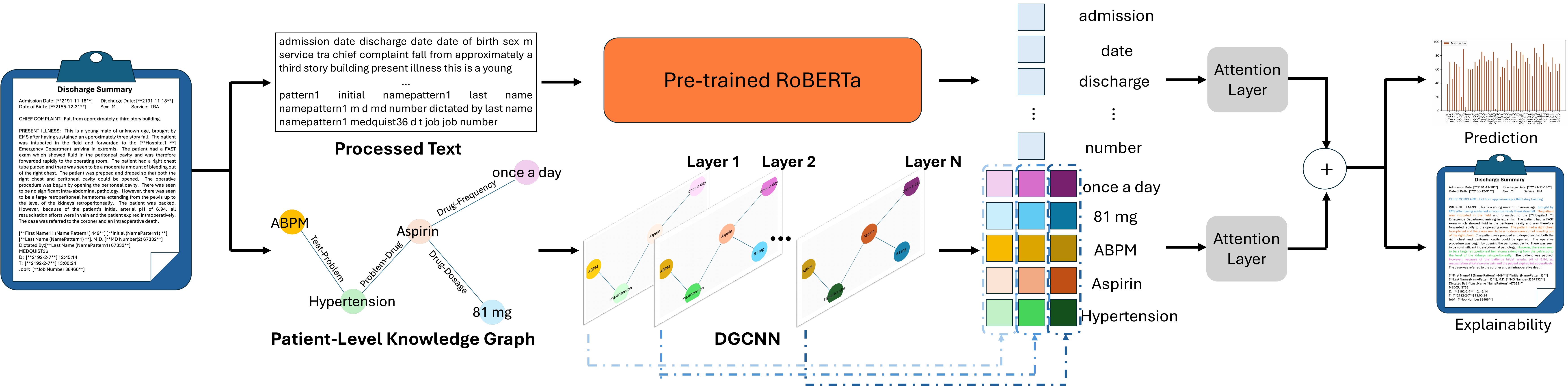} 
\caption{Architecture of the proposed model. The processed discharge summary as input is encoded using a pre-trained RoBERTa, while its corresponding patient-level knowledge graph inputs a DGCNN module, with final representations obtained by concatenating node features from all layers. Both representations are fed into separate label-wise attention layers, after which the weighted outputs are concatenated, using for ICD code prediction. 
}
\label{fig2}
\end{figure*}


In this section, we detail the construction of patient-level knowledge graphs and their integration into the PLM-ICD coding architecture.

\paragraph{Patient-Level Knowledge Graph Construction}

We aim to construct patient-level knowledge graphs that comprehensively represent a patient’s medical history, encompassing diseases, treatments, tests, drugs, dosages, frequencies, strengths, and so on, as well as the relationships between these entities. We employ named-entity recognition (NER) and relation extraction (RE) models provided by Healthcare NLP library \cite{johnsnowlabs2024} to extract these concepts.

Out of the available RE models in Healthcare NLP, 
we select five models based on the quantity of triples extracted and their uniformity 
across all documents. The selected RE models are (ordered by frequency) \emph{`Clinical Relationship'} (CR), \emph{`Temporal Events'} (TE), \emph{`Posology Relationship'} (PR), \emph{`Bodypart-Directions'} (BD) and \emph{`Bodypart-Problem'} (BP). These models collectively identify 14 different types of entities. Detailed information about model selection, selected RE models and statistics of the extracted entities and relationships can be found in 
in Appendix A.1.

The output of these relationship extraction (RE) models includes two identified entities, their respective types, and the relationship between them. 
When constructing a patient’s knowledge graph, we represent this information as triples in the format $<entity1,relationship,entity2>$ (e.g., \mbox{$<lisinopril,drug$-$strength,40mg>$}). 
The resulting patient-level knowledge graphs contain four types of information (For a visualisation, consult Appendix A.2):

\textbf{Diagnostic Information (CR):} Revealing the interrelationships among problems, treatments, and tests;

\textbf{Temporal Information (TE):} Capturing the sequence of clinical events;

\textbf{Posology Information (PR):}  Providing details on drug regimens, including dosage, duration, strength, and frequency, as well as their interrelationships;

\textbf{Anatomical Information (BD and BP):} Illustrating the connections between problem or directions and specific body parts.

The statistics of the graphs extracted from the two MIMIC-III datasets, Full and Top-50, are summarized in Table~\ref{tab:statistics of graph}. On average, the graphs contain approximately 190 nodes, with each node typically comprising around two tokens. The largest graph in the dataset includes 903 nodes, while some documents don't have any extracted graphs.

Furthermore, we evaluate the quality of the constructed graphs from a \textbf{\textit{Statistical Perspective}} by measuring information loss. Specifically, we calculate the average information entropy of the original text and the serialized graph. As shown in Table~\ref{tab:entropy whole}, our analysis indicates that the extracted content accounts for less than 23\% of the original size, yet retains approximately 90\% of the information. This highlights the efficiency of our patient-level knowledge graph in significantly compressing the text while preserving the majority of its informational content. (For details of the information entropy methodology and further results of the ablation study, conducted by removing each type of entity and relationship, please refer to Appendix A.3.)

\begin{table}
    \centering
    \resizebox{1\columnwidth}{!}{%
    \begin{tabular}{cccccc}
    \hline
     \textbf{Split} & \textbf{Avg  $|T|$}&   \textbf{Avg $|N|$}&   \textbf{Avg $T$ in $N$} &\textbf{Min/Max $T$} &\textbf{Min/Max $N$}\\
    \hline
    Full & 1513.5&  183.0& 342.3 & 0/1954 & 0/903  \\
    Top-50 & 1612.0&  196.8& 366.8 & 6/1689 & 3/774 \\
    \hline
    \end{tabular}
    }
    \caption{Statistics of nodes and tokens per processed document in MIMIC-III datasets. $T$ stands for tokens, $N$ stands for graph nodes. \emph{`Avg'} represents averages over all documents.}
    \label{tab:statistics of graph}
\end{table}
\begin{table}
    \centering
    \resizebox{1\columnwidth}{!}{%
    \begin{tabular}{ccccc}
        \hline
        \textbf{Dataset} & \textbf{Text Entropy} & \textbf{Graph Entropy} & \textbf{Ratio (\%)} \\
        \hline
        Full & 8.33 & 7.48 &  89.95 \\
        Top-50 & 8.41 & 7.61 &  90.52 \\
        \hline
    \end{tabular}
    }
    \caption{The Information entropy of processed text and serialised graph. The \emph{`Ratio'} measures how much information is retained.}
    \label{tab:entropy whole}
\end{table}




\paragraph{Task Definition}



ICD coding is formulated as a multi-label classification task. Given a clinical document (discharge summary in MIMIC-III) of a patient, automated coding module aims to assign the correct ICD codes which represent the diseases or procedures. Specifically, we define a clinical document with $N_t$ tokens as $\mathbf{d} = \{t_1,t_2,...,t_{N_t}\}$. The goal is to predict a distribution of labels $\mathbf{p} = \{p_1,p_2,...,p_{N_c}\}$, where $N_c$ denotes the total number of codes in the label space. The final set of assigned codes is the ones that exceed a pre-defined probability threshold.


The proposed framework is shown in Figure~\ref{fig2}. 
 The subsequent sections will provide a detailed description of each component of the framework.

\paragraph{Text Embedding - Pre-trained Language Model}
To embed the textual data, we utilize RoBERTa-PM \cite{lewis2020pretrained}, a transformer model pre-trained on biomedical abstract and clinical documents.

The pre-processing of the raw text in MIMIC-III datasets follows~\citet{mullenbach2018explainable}. Following PLM-ICD, we divide each document 
into segments of equal length of $l$ tokens. The number of segments per document is represented as $N_s$ and varies across different samples. Thus, each segment comprises a sequence of tokens that represent a portion of the document:
\begin{equation}
    s_i = \{t_j|l \cdot i \leq j < l \cdot (i+1)\}.
\end{equation}
The document representation $\mathbf{H}_{\textmd{t}}$ is formed by concatenating the hidden representations of each segment:
\begin{equation}
    \mathbf{H}_{\textmd{t}} = \textmd{concat}(PLM(s_1),...,PLM(s_{N_s})),
\end{equation}
where $PLM(s_i)$ denotes the representation for segment $s_i$ embedded by RoBERTa-PM.
\paragraph{Graph Embedding - Deep Graph Convolutional Neural Network}
The Deep Graph Convolutional Neural Network (DGCNN)~\cite{zhang2018end} we refer to in this work is an end-to-end architecture designed for graph classification tasks. 
But we represent the graph using the hidden state from the final layer of DGCNN, just before the SortPooling layer in the original framework, as this configuration is found to yield the best performance based on initial experimental results.

Given a patient's knowledge graph $G$, we can obtain its adjacency matrix $\mathbf{A}$ and diagonal degree matrix $\mathbf{D}$. 
The hidden state of the first graph convolution layer is as follows:
\begin{equation}
    \mathbf{H}_{\textmd{g}}^1 = f(\mathbf{D}^{-1}\mathbf{A}\mathbf{X}\mathbf{W}),
\end{equation}
where $\mathbf{X} \in \mathbb{R}^{N_n \times d_n}$ denotes the node representation matrix with dimension $d_n$; $N_n$ represents the number of nodes in the graph; $\mathbf{W} \in \mathbb{R}^{d_n\times d_n'}$ is a trainable parameter matrix, in which $d_n'$ defines the dimension of code representation for the next convolution layer; $f$ is a nonlinear activation function.

DGCNN adopts multiple convolution layers, as it allows for the extraction of multi-scale local substructure features. Therefore, the output of the $m^{th}$ graph convolution layer is represented as follows:
\begin{equation}
    \mathbf{H}_{\textmd{g}}^{m+1} = f(\mathbf{D}^{-1}\mathbf{A}\mathbf{H}_{\textmd{g}}^m\mathbf{W}^m),
\end{equation}
where $\mathbf{H}_{\textmd{g}}^0 = \mathbf{X}$. The final representation of patient's knowledge graph $\mathbf{H}_{\textmd{g}}$ is the concatenation of the features from all $[\mathbf{H}_{\textmd{g}}^1,...,\mathbf{H}_{\textmd{g}}^{N_y}]$,
where $N_y$ is the number of graph convolution layers.
\paragraph{Multi-Head Label-Wise Attention}
To capture label-specific information and assign varying attention weights to fragments (tokens or nodes) for each label, we incorporate a label-wise attention layer following the patient representation. Instead of just feeding the concatenated representation of text $\mathbf{H}_{\textmd{t}}$ and graph $\mathbf{H}_{\textmd{g}}$ to a single attention layer, we utilize a multi-head attention mechanism. This approach enables the model to focus on information from different representation sub-spaces. Consequently, $\mathbf{H}_{\textmd{t}}$ and  $\mathbf{H}_{\textmd{g}}$ are processed through separate label-wise attention layers. The attention score matrices are defined as follows:
\begin{equation}
    \alpha_{\textmd{t}} = \textmd{softmax}(\mathbf{V}_1\tanh(\mathbf{V}_2\mathbf{H}_{\textmd{t}}),
   \end{equation}
    \begin{equation}
    \alpha_{\textmd{g}} = \textmd{softmax}(\mathbf{V}_3\tanh(\mathbf{V}_4\mathbf{H}_{\textmd{g}}),
\end{equation}
where $\mathbf{V}_{1-4}$  are trainable linear transformation matrices. The weighted label-specific representations are calculated as follows:
\begin{equation}
    \mathbf{Z}_{\textmd{t}} = \mathbf{H}_{\textmd{t}}\alpha_{\textmd{t}}^T,
    \mathbf{Z}_{\textmd{g}} = \mathbf{H}_{\textmd{g}}\alpha_{\textmd{g}}^T.
\end{equation}
Finally we concatenate them to form a representation for the individual patient $\mathbf{Z} = [\mathbf{Z}_{\textmd{t}},\mathbf{Z}_{\textmd{g}}]$.
The probability of predicting label $i$ is calculated by:
\begin{equation}
    \mathbf{p}_i = \sigma(\mathbf{L}_i \cdot \mathbf{Z}_i),
\end{equation}
where $\mathbf{L}_i$ is the representation of the $i^{th}$ label and $\mathbf{Z}_i$ is the label-specific patient representation.
The final predicted soft-maxed probability vector $\hat{\mathbf{y}}$ and true labels $\mathbf{y}$ are used to compute the binary cross-entropy loss:
\begin{equation}
    \mathcal{L}(\mathbf{y},\mathbf{p})=-\frac{1}{|\mathbf{y}|}\sum_{i=1}^{|\mathbf{y}|}\left(\mathbf{y}_i\log\hat{\mathbf{y}}_i+(1-\mathbf{y}_i)\log(1-\hat{\mathbf{y}}_i)\right).
\end{equation}
\begin{table*}[t]
    \centering
    \resizebox{2\columnwidth}{!}{%
    \begin{tabular}{lccccccccccc}
    \hline
    & \multicolumn{5}{c}{\textbf{MIMIC-III Full}} & & \multicolumn{5}{c}{\textbf{MIMIC-III Top-50}}\\
    \cline{2-6}
    \cline{8-12}
    & \multicolumn{2}{c}{F1} & \multicolumn{2}{c}{AUC} & Precision & & \multicolumn{2}{c}{F1} & \multicolumn{2}{c}{AUC} & Precision  \\ 
    
    Model & Macro & Micro & Macro & Micro & P@8 & & Macro & Micro & Macro & Micro & P@5 \\
        \hline
        MultiResCNN & 9.0 & 55.2 & 91.0 & 98.6 & 73.4 & & 59.29 & 66.24 & 89.30 & 92.04 & 61.56 \\
        2Stage & 10.5 & 58.4 & 94.6 & 99.0 & 74.4 & & \textbf{68.93} & \textbf{71.83} & \textbf{92.58} & \textbf{94.52} & 66.72 \\
        JointLAAT & 10.2 & 57.5 & 92.1 & 98.8 & 73.5 & & 66.95 & 70.84 & 92.36 & 94.24 & 66.36 \\
        MSMN & 10.24 & 58.70 & \textbf{94.78} & \textbf{99.15} & 75.45 & & 66.68 & 71.19 & 92.12 & 94.21 & 66.86 \\
        \hline
        PLM-ICD & 9.69 & 59.06 & 92.12 & 98.83 & \textbf{76.72} & & 64.61 & 70.33 & 91.16 & 93.63 & 66.11\\
        Our Model & \textbf{11.05} & \textbf{59.72} & 92.37 & 98.75 & 76.59 & & 67.81 & 71.63 & 92.04 & 94.22 & \textbf{67.08}\\
        \hline
    \end{tabular}
    }
    \caption{Results on the MIMIC-III Full and Top-50 test sets. The best results are highlighted in bold.}
    \label{tab:ICD coding result}
\end{table*}

\section{Empirical Evaluation}
\subsection{Experiment Setup}




\paragraph{Datasets and Metrics}
Like most evaluation methods for multi-label classification tasks, clinical coding is typically assessed using three standard metrics: F1, AUC and Precision@N. In this work, we utilize these metrics to evaluate the models on two commonly used datasets: MIMIC-III Full and MIMIC-III Top-50.

MIMIC-III is a publicly accessible database comprising de-identified health data from patients admitted to critical care units at the Beth Israel Deaconess Medical Center in Boston, Massachusetts between 2001 and 2012. 
The standard clinical coding task involves using discharge summaries from the MIMIC-III dataset to assign ICD-9 codes, which include discharge diagnoses and procedures. 

The MIMIC-III Full dataset includes 52,723 documents from 41,126 patients, with each document containing a median of 1,375 words and 14 codes. The MIMIC-III Top-50 dataset focuses on the top 50 most frequent diagnosis and procedure codes from the Full dataset. It consists of 11,368 documents from 10,356 patients, with a median of 1,478 words and 5 codes per document.
\paragraph{Implementation Details}
We train our model using four 80GB NVIDIA A100 GPUs within an environment configured with CUDA 11.1 and PyTorch 1.12.0. 
Detailed implementation hyperparameters for both our model and PLM-ICD are provided in Appendix A.4.
\paragraph{Baselines}
To demonstrate the effectiveness of our model, we compare it with five current SOTA approaches.\\
\textbf{PLM-ICD} \cite{huang2022plm}, leverages transformer-based models specifically pre-trained on biomedical and clinical texts. It achieves SOTA performance on both MIMIC-III and MIMIC-IV datasets \cite{edin2023automated}. We select it as our base model due to its strong performance as a widely used baseline and its simple structure, which facilitates the integration with the graph representation module.\\
\textbf{MultiResCNN} \cite{li2020icd} employs a multi-filter convolutional layer to capture text patterns of varying lengths and a residual convolutional layer to expand the receptive field.\\
\textbf{2Stage} \cite{nguyen2023two} leverages the hierarchical properties of codes to perform predictions in two sequential steps.\\
\textbf{JointLAAT} \cite{vu2020label} introduces a hierarchical joint learning mechanism to address label imbalance.\\
\textbf{MSMN} \cite{yuan2022code} utilizes synonyms with multi-head attention mechanism, achieving another state-of-the-art performance on MIMIC-III Full.
\begin{figure}[t]
    \centering
    \begin{tabular}{cc}
        \includegraphics[width=0.22\textwidth]{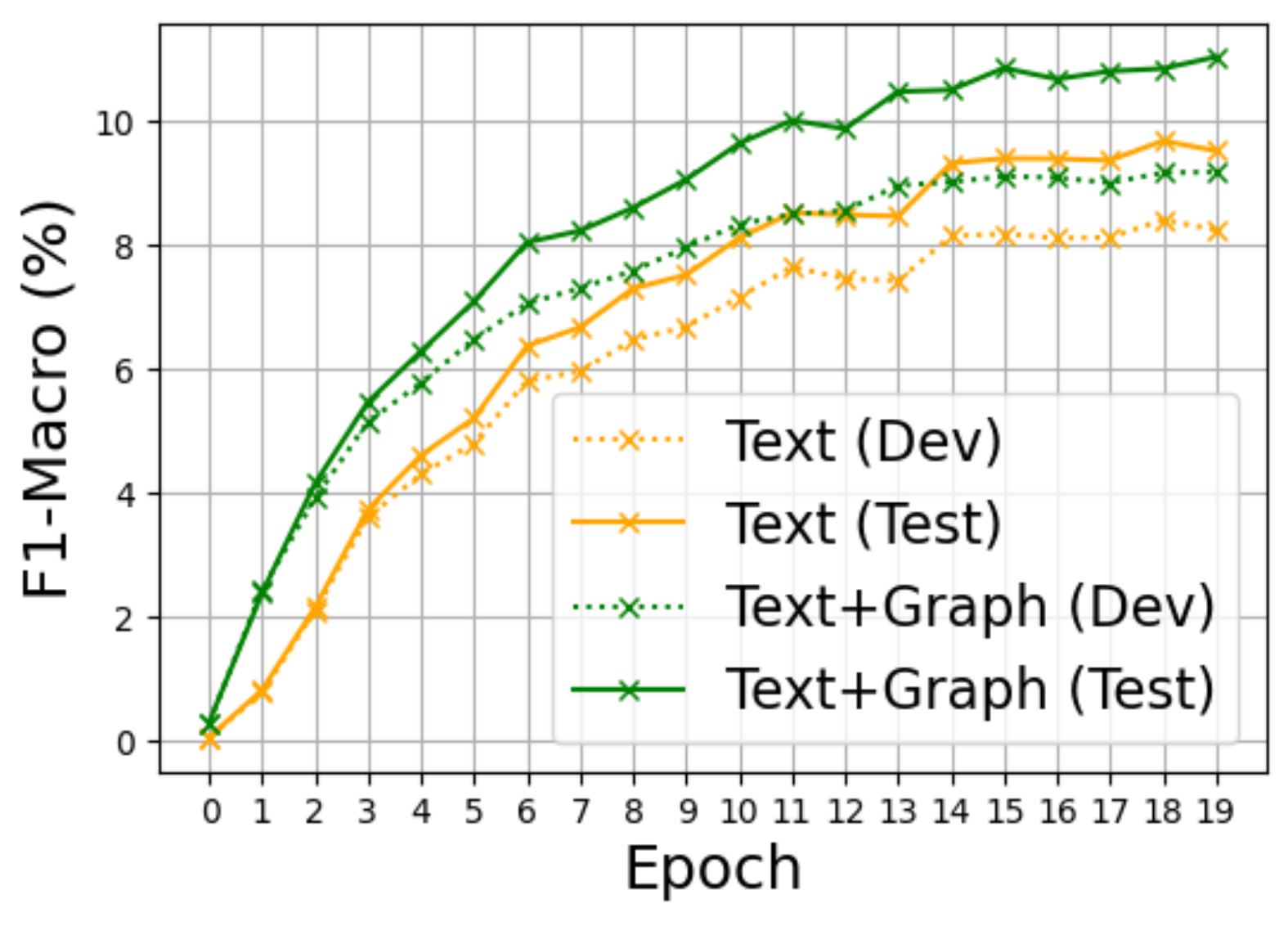} &
        \includegraphics[width=0.22\textwidth]{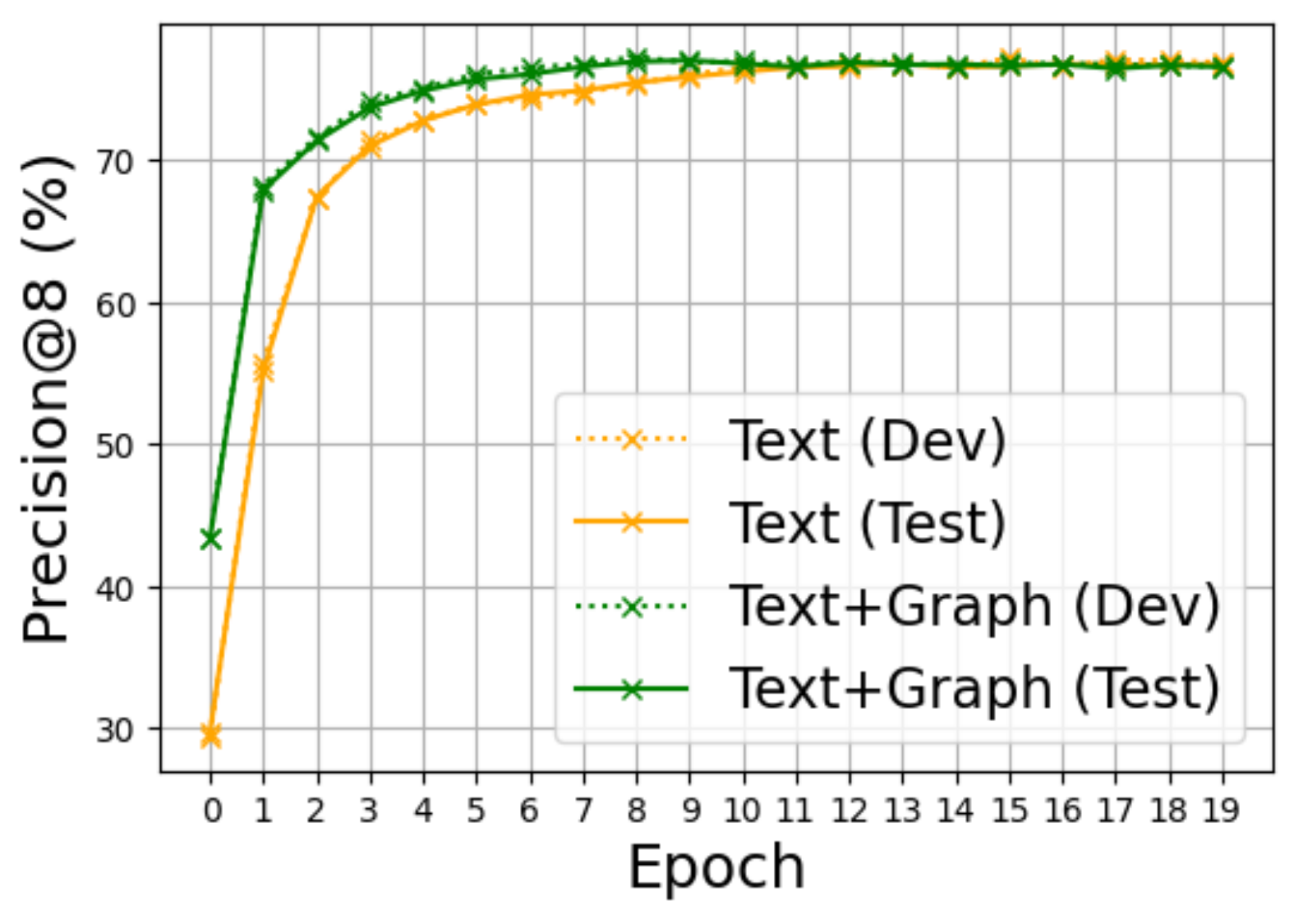} \\
        \includegraphics[width=0.22\textwidth]{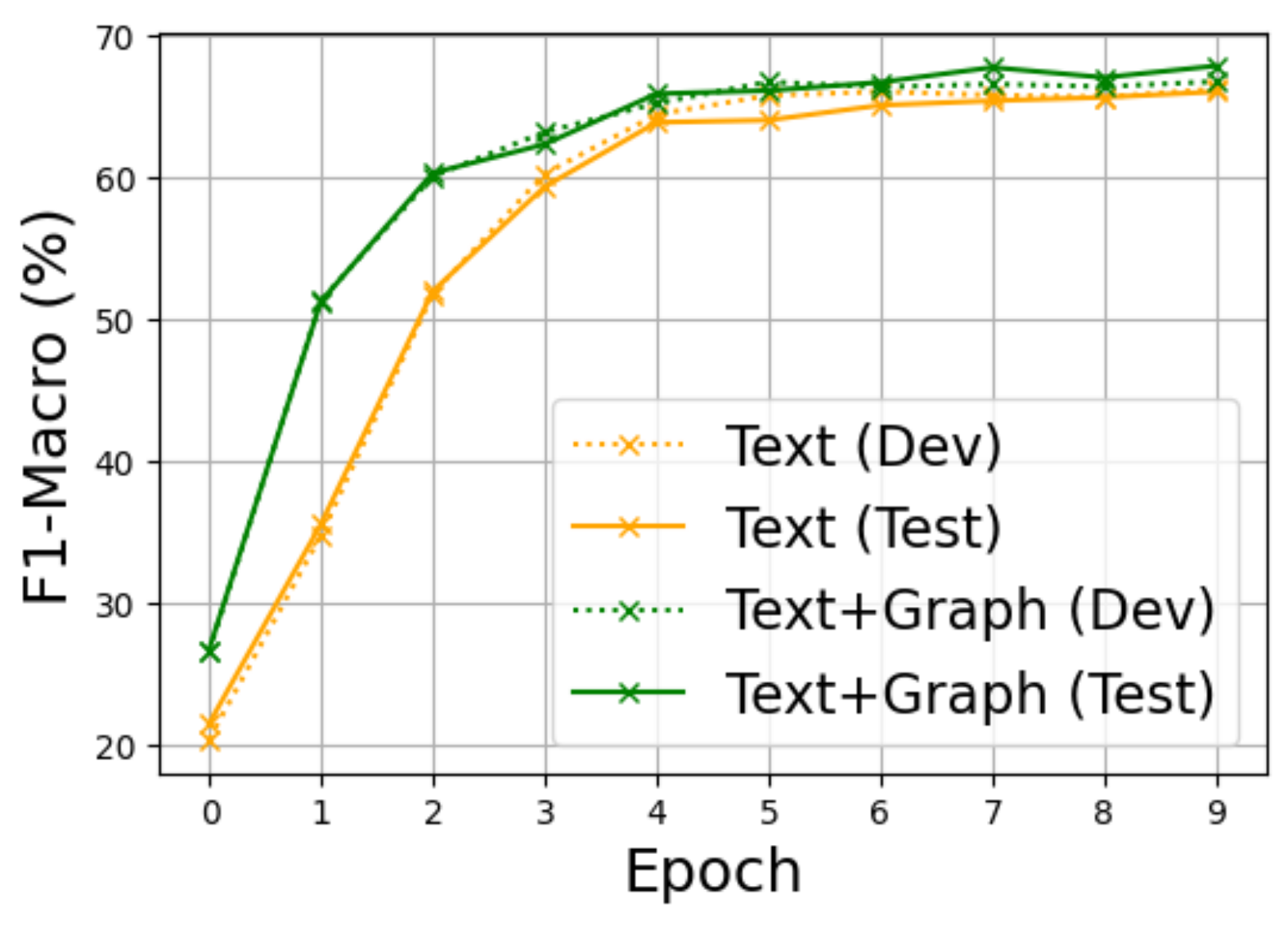} &
        \includegraphics[width=0.22\textwidth]{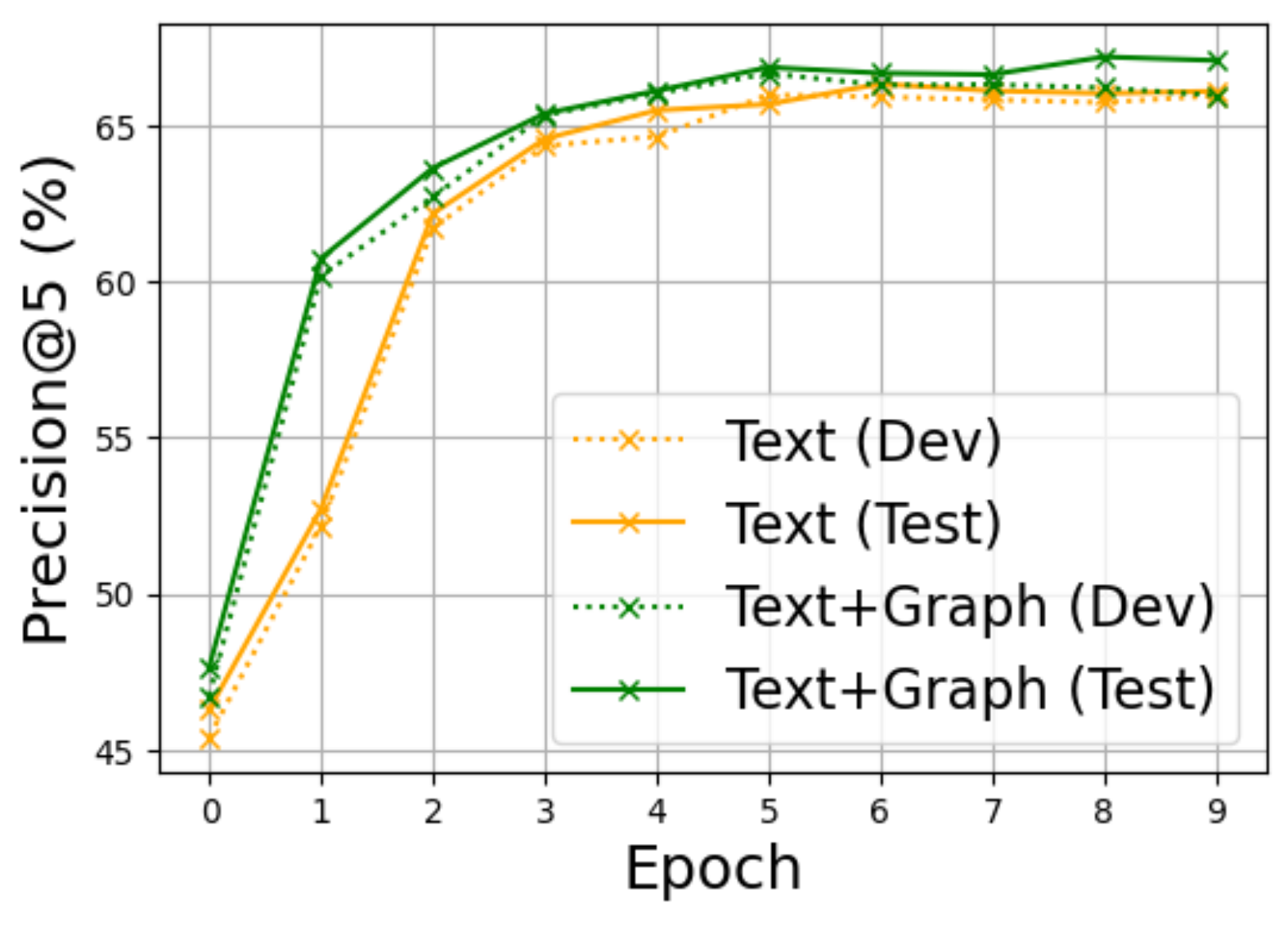} \\
    \end{tabular}
    \caption{By-epoch performance comparison of our model and PLM-ICD by means of Macro-F1 / P@8 on MIMIC-III Full (top row) and Macro-F1 / P@5 on MIMIC-III Top-50 (bottom row).}
    \label{fig:converge}
\end{figure}
\begin{table}[!t]
\centering
\resizebox{0.95\columnwidth}{!}{%
    \begin{tabular}{lcccccc}
        \toprule
        \hline
        \textbf{Remove} & \textbf{$\mu$F1} & \textbf{$m$F1} & \textbf{$\mu$AUC} & \textbf{$m$AUC} & \textbf{P@8} \\ 
        \hline
        \midrule
        Full & 11.05 & 59.72 & 92.37 & 98.75 & 76.59 \\ 
        \hline
        \midrule
        $-$BP & 10.38 & 59.60 & 92.39 & 98.86 & 76.72 \\ 
        $-$PR & 10.33 & 59.65 & 92.39 & 98.84 & 76.95 \\
        $-$TE & 10.34 & 59.45 & 92.53 & 98.86 & 76.62 \\
        $-$CR & 10.07 & 59.35 & 92.63 & 98.89 & 76.74 \\
        $-$BD & 10.61 & 59.51 & 92.24 & 98.79 & 76.51 \\
        \midrule
        \hline
        $-$drug & 10.52 & 59.44 & 92.23 & 98.77 & 76.61 \\
        $-$problem & 9.77 & 59.26 & 92.35 & 98.86 & 76.95 \\
        $-$treatment & 10.76 & 59.66 & 92.33 & 98.81 & 76.76 \\
        $-$test & 10.72 & 59.59 & 92.32 & 98.81 & 76.54 \\
        \hline
        \bottomrule
    \end{tabular}
}
    \caption{Results of ablation study on the MIMIC-III Full dataset. Removing all relationships and entities of a specified type. $\mu$ and $m$ denote Macro and Micro averages, respectively.}
    \label{tab:B-2}
\end{table}
\subsection{Quantitative Results}
\paragraph{A. Does integrating graph-based representation enhance the ICD coding performance?}
This experiment aims to verify if integrating the patient-level knowledge graph benefits the representation of the patient, consequently enhances the performance of ICD coding. 
The results shown in Table \ref{tab:ICD coding result} indicate that our model outperforms its base model PLM-ICD significantly on the F1-Macro score by 1.36\% and 3.20\% on the Full and Top-50 datasets, respectively. F1-Macro score is the primary metric for this task due to its effectiveness in balancing precision and recall across classes and its robustness in classification problems. Our model exhibits more noticeable performance improvements on frequent labels and demonstrates overall advancements across all metrics. Moreover, our model remains highly competitive compared to other state-of-the-art methods, achieving the highest F1 scores on full label set. 

Additionally, our model achieves higher scores in the early epochs (see Figure \ref{fig:converge}), highlighting its efficiency when computational resources are constrained. The most significant improvements occur within the first three epochs, indicating that the structured information is efficiently captured early. These findings further validate the quality of the constructed graphs, demonstrating their effectiveness in patient representation (\textbf{\textit{Statistical Perspective}}) by providing not only semantic information but also additional structured information.
\paragraph{B. What elements should constitute a patient’s knowledge graph?}
\paragraph{Relationship}
We conduct an ablation study to assess the impact of different types of relationships in the graph on patient representation. 
By removing a single type of relationship from the complete graph, we observe that the removal of any relationship leads to a noticeable decrease in performance.
Despite this, the performance still remains superior to the base model PLM-ICD by at least 0.4\% on F1-Macro score. Excluding the \emph{`Clinical Relationship'} (CR) results in the most substantial drop in performance, indicating its critical importance in patient representation. 
From Table 5 in Appendix A we can see that the number of \emph{`Clinical Relationships'} (CR) is similar to \emph{`Temporal Events'} (TE) in MIMIC-III Full dataset. But its exclusion causes a more pronounced decline, suggesting that its significance lies not only in its quantity but also in the quality of information it provides about the patient. This is intuitive, as \emph{`Clinical Relationships'} (CR) inherently capture the essential aspects of a patient's profile—such as medical problems, treatments, and diagnostic tests—that are directly relevant to predicting diseases and procedures codes. 
Conversely, \emph{`Bodypart-Directions'} (BD) has the least impact on ICD coding. 
\paragraph{Entity}
We conduct another ablation study by removing entities of the four most occurring types: \emph{`Problem'}, \emph{`Test'}, \emph{`Treatment'}, and \emph{`Drug'} (ordered by frequency). The removal of \emph{`Problem'} has the most significant impact on the F1-Macro score, indicating that \emph{`Problem'} plays a crucial role in the graph representation. This finding also make sense intuitively, as \emph{`Problem'} constitutes the largest portion of the graph and is most closely related to the objective of diagnosing the patient.
\subsection{Qualitative Results}
\begin{figure}[t]
\centering
\includegraphics[width=0.43\textwidth]{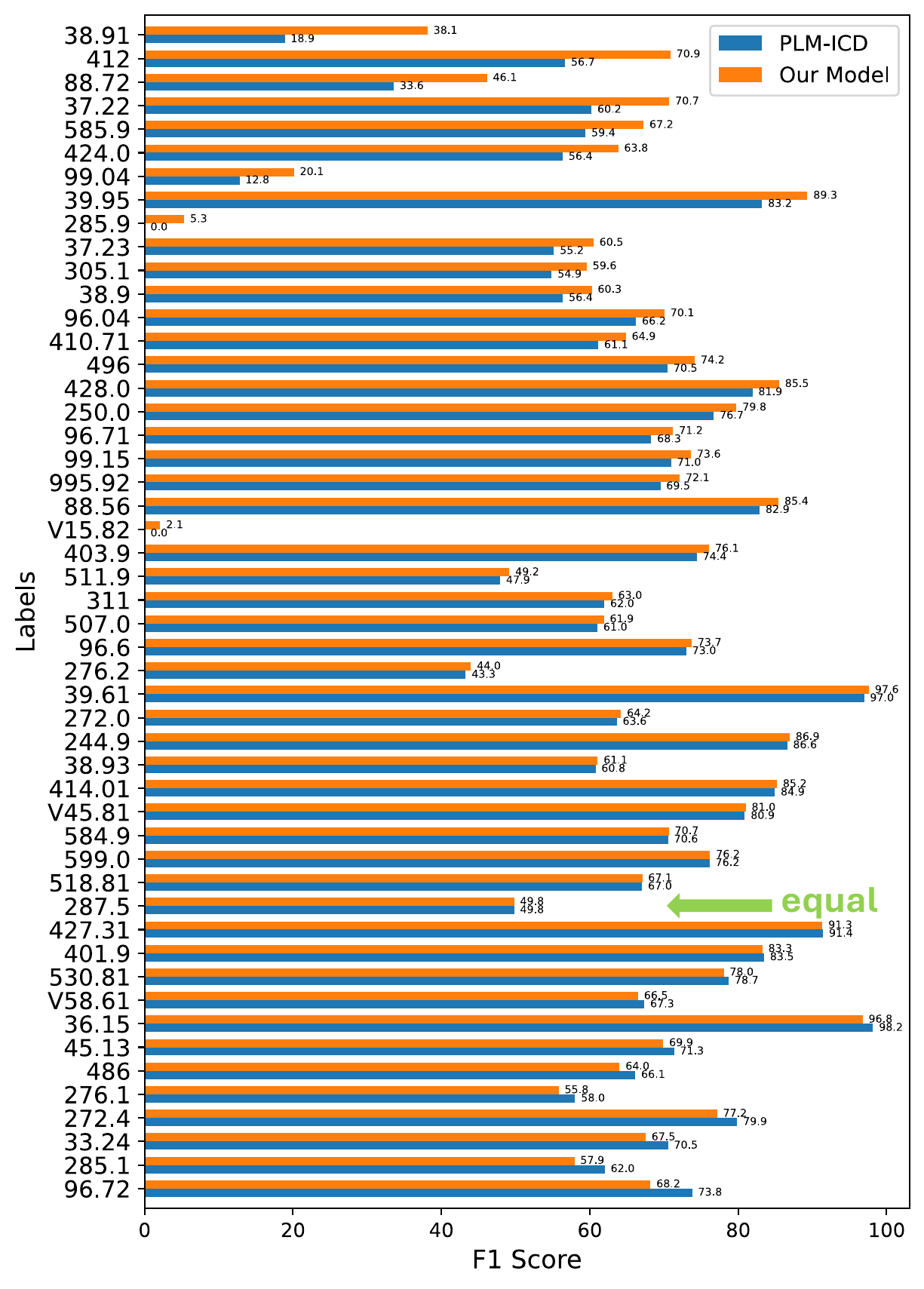} 
\caption{F1 performance comparison on each of the top-50 codes between our model and PLM-ICD, ranked by the performance difference between the two models.}
\label{category}
\end{figure}

\paragraph{C. How does the patient-level knowledge graph help the classification for specific codes?}
To further analyse performance at the label level, we compute the F1 scores for our model and PLM-ICD on the MIMIC-III Top-50 dataset for each code (see Figure~\ref{category}). The results reveal that our model outperforms PLM-ICD on 37 codes out of 50. Notably, our model achieves scores for codes \texttt{285.9} (\emph{Anemia, unspecified'}) and \texttt{V15.82} (\emph{Personal history of tobacco use'}), which PLM-ICD totally fails.
\begin{figure}[t]
    \centering
    \includegraphics[width=\columnwidth]{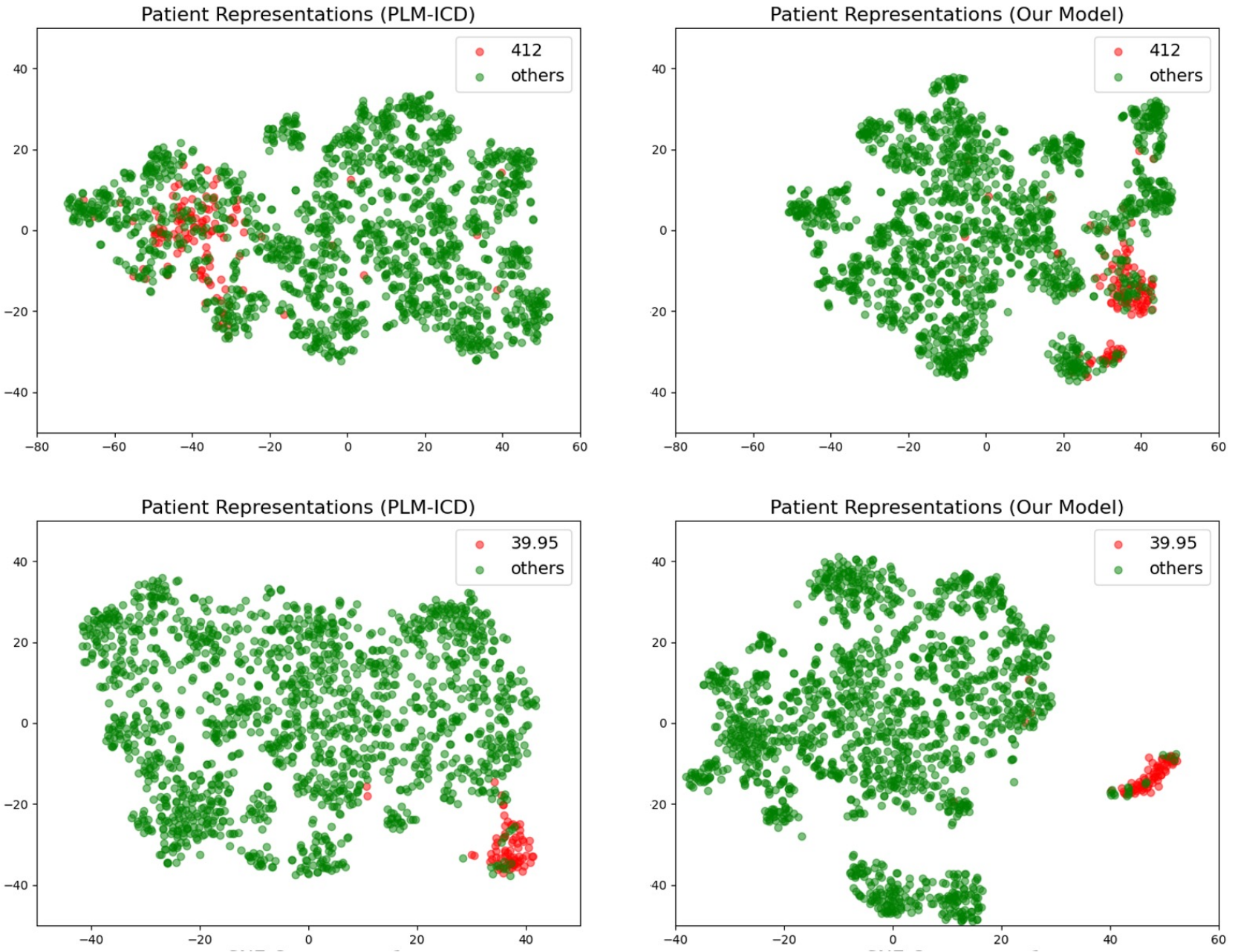}
    \caption{Visualisation of label-specific patients representation of codes \texttt{412} \emph{`Old myocardial infarction'} and \texttt{39.95} \emph{`Hemodialysis'}, without (left) and with (right) using knowledge graphs as input. Instances with the corresponding ground-truth label are \textcolor{red}{red}.}
    \label{165048}
\end{figure}
To better understand how graphs enhance patient representations, we visualize the label-specific representations of all samples in the test set (see Figure~\ref{165048}). We focus on the codes \texttt{412} (\emph{Old myocardial infarction'}) and \texttt{39.95} (\emph{Hemodialysis'}) (see Appendix A.5), where both our model and PLM-ICD demonstrate good performance. This choice avoids complications from low scores, which may result in erratic embeddings that are challenging to visualize, such as the case of \texttt{38.91} \emph{`Arterial Catheterization'}. Samples with the corresponding labels are highlighted in red. Specifically, we reduce the dimensionality of the original representations $\mathbf{Z_i}$ using t-SNE. For code \texttt{412}, our model exhibits a noticeably higher density of instances with the target label (red), with an average distance of 16.44 between positive points compared to 19.14 for PLM-ICD. For code \texttt{39.95}, where both models perform well, our model still shows a denser cluster of the positive (red) instances, and the cluster is more distinctly separated from other points. This case study demonstrates that integrating structured information enhances patient representation, leading to more accurate classification.
\begin{figure}[t]
    \centering
    \includegraphics[width=\columnwidth]{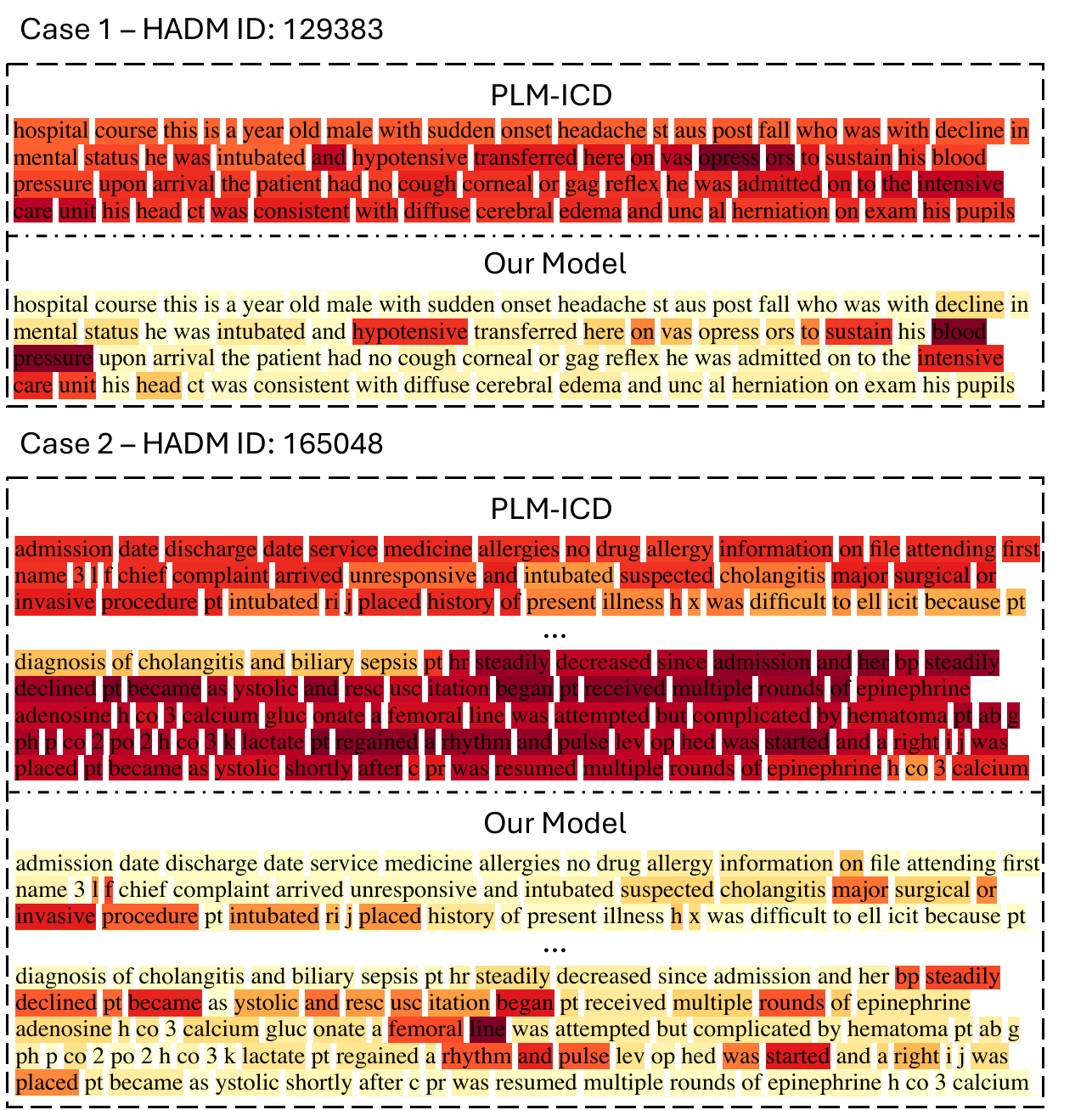}
    \caption{Highlights related to label \texttt{38.91} \emph{`Arterial Catheterization'}, without (above) and with (below) using knowledge graphs as input.}
    \label{highlight}
\end{figure}
\paragraph{D. Explainability}
The ability to provide trustworthy and interpretable explanations is particularly critical in the clinical domain. To achieve this, we highlight text spans based on their attention weights, using darker colors to indicate higher weights. This suggests that these spans contribute more significantly to representing the patient. Our model demonstrates the ability to identify the most relevant spans more accurately and concisely. To illustrate this, we present two non-cherry-picked examples from the test set on label \texttt{38.91}: \emph{`Arterial Catheterization'}, where our model shows the most improvement. In Case 1 (Figure \ref{highlight}, above), our model effectively captures key tokens like \emph{`hypotensive'} and \emph{`blood pressure'}, which are directly associated with \emph{`Arterial Catheterization'}, whose role is continuous blood pressure monitoring and arterial blood gas analysis. In contrast, PLM-ICD distributes attention more evenly across the text. In Case 2 (Figure \ref{highlight}, below), our model successfully highlights relevant spans across various sections, such as \emph{`invasive procedure'} and \emph{`placing a femoral line'}, they are procedures often involved in \emph{`Arterial Catheterization'}. Additionally, phrases like \emph{`intubated rij placed'} and \emph{`a right IJ was placed'} are highlighted as they pertain to \emph{`central venous catheterization'}, which is another type of catheterization. The model also succinctly highlights \emph{`rhythm and pulse'}, which is related to blood pressure monitoring. These two cases strongly demonstrate that our model excels in providing high-quality explanations compared to PLM-ICD.

\section{Conclusion}
In this work, we construct a patient-level knowledge graph comprising wide range of entities and relationships. We integrate it into a state-of-the-art ICD coding architecture, PLM-ICD, which significantly enhances the patient representation and improve the coding performance. Additionally, we verify the impact of different types of entities and relationships in representing the patient. Furthermore, we showcase how integrating graph improves the patient representation through visualisation and demonstrate the high-quality explainability of our model in case studies.

Our patient-level knowledge graph dataset holds significant potential to provide healthcare providers with more precise, data-driven insights, ultimately improving patient outcomes, such as optimizing treatment plans and enabling early diagnosis.
\clearpage
\section{Limitations}

While our model demonstrates the effectiveness of integrating graph-based information, more advanced models have been developed in parallel (e.g., \cite{nguyen2023two}). Our results confirm the benefits of structured knowledge integration by comparing our model to its baseline, highlighting the quality of our patient KG. However, we acknowledge that integrating structured information into these more advanced models could further enhance their performance, which remains an avenue for future exploration.

This work serves as the evaluation of the constructed patient-level KG from two perspectives: statistical and representational. However, it lacks a comprehensive comparison with other types of patient KGs (except for \citet{yuan2021graph}) due to the challenges and time constraints associated with constructing them based on MIMIC-III.

In future work, we will also aim to enrich the patient-level knowledge graph by integrating other knowledge sources, such as hierarchical information from ontology systems like SNOMED-CT and UMLS. In the current study, we did not account for the semantic meaning of edges within graph representation, as some links merely signify connections between entities (e.g., \emph{`1'} or \emph{`TREATMENT-TEST'}). Moving forward, we plan to model the meaning of these relationships more explicitly by combining their semantic representations with confidence measurements. \

Additionally, we have not  explored other advanced graph representation models, such as Relational Graph Convolutional Networks (R-GCN)~\cite{schlichtkrull2018modeling} and Graph Attention Networks (GAT)~\cite{velivckovic2017graph}. The application of GAT, in particular, offers potential for further enhancing explainability by identifying and highlighting the sub-graphs that contribute most to final predictions, which we aim to evaluate more rigorously in domain expert-centred experiments.

Finally, due to resource constraints, we have not experimented with adapting other baseline models to use the document-level structured representation graphs. It is unlikely, but not impossible, that other architectures would not benefit from this kind of information, and further experiments should be conducted to establish this fact empirically.

\bibliography{acl_katex}

\appendix
\section{Appendix}
\begin{table*}[h]
    \centering
    \begin{tabular}{ccc}
    \hline
    \textbf{RE Model}&   $|Tr|$ & $|S|$\\
    \hline
    clinical relatioship&  6878467 & 52721\\
     temporal events &  6504349 & 52720\\
     posology relationship & 3939341 & 51879\\
     ade conversational relationship& 2443125 & 12464\\
    bodypart-directions & 355260 & 42487\\
     bodypart-problem & 337041 & 38719\\
     ade relationship& 86062 & 24259\\
     test-problem-finding & 76262 & 29007\\
     drugprot relationship & 42071 & 16859\\
     bodypart-proceduretest&  14739 & 8861\\
     generic relationship &7004 & 2897\\
    date relationship& 2979 & 1713\\
    test-result-date & 2174 & 2174\\
    phenotype gene relationship& 0 & 0\\
    \hline
    \end{tabular}
    \caption{Statistics of RE model outputs in the MIMIC-III Full dataset. \emph{$|Tr|$} refers to the number of triples recognized by the RE model. \emph{$|S|$} indicates the number of samples in the full dataset that contain these triples.}
    \label{tab:statistics1}
\end{table*}
\subsection{Patient-Level Knowledge Graph Construction}
\paragraph{Model Selection}
The Healthcare NLP library includes 44 RE models, each integrating both NER and RE functionalities. These models are trained on various language models across multiple languages to extract a wide range of clinical information. We utilize 14 of these models, which cover all available relationship types except for \emph{`drug-drug interaction'} and share a consistent architecture. Details and statistics of these RE models are provided in Table \ref{tab:statistics1}.

The top five relationships include \emph{ade conversational'}, which links drugs to their adverse reactions. However, we do not select it due to its uneven distribution across samples, as only a limited number contain this type of triple. Instead, we chose the \emph{`bodypart-problem'} relationship, which ranks sixth.
\paragraph{Selected 5 RE Models}
Table \ref{tab:RE_models} details the entities and relationships that each RE model can extract. The entities recognized from the MIMIC-III notes include a subset of those listed in this table.
\paragraph{Statistics of Entities Extracted}
The complete patient-level knowledge graph, which encompasses all five relationships, identifies 14 types of entities. The statistics of them can be found in Table \ref{tab:statistics_entities1}. In the ablation study, we study the impact of top four types of entities, as they have the highest magnitude compared to others.
\begin{table*}[h]
    \centering
    \begin{tabular}{cc}
    \hline
    \textbf{Entity Type}&   $|En|$\\
    \hline
     problem&  3422556\\
     treatment& 1665523\\
     test&  1371889\\
     drug& 1039115\\
     strength& 636491\\
     frequency& 338332\\
     form & 229420\\
     dosage & 217178\\
     internal organ or component &192503\\
    route& 166454\\
    direction & 135903\\
    symptom& 106114\\
    external body part or region& 86367\\
    duration& 41727\\
    \hline
    \end{tabular}
    \caption{Statistics of entities identified in the MIMIC-III Full dataset. $|En|$ represents the number of entities.}
    \label{tab:statistics_entities1}
\end{table*}
\subsection{Patient-Level Knowledge Graph Visualisation}
\label{sec:Visualisation}
Figure \ref{fig:visualisation} presents a visualization of a patient-level knowledge graph (HADM ID: 196292). To make it clear, we include type information for the entities, linking each entity to its respective type. Nodes representing types are colored light green, while different types of entities assigned unique colors.

\subsection{Methodology of Information Entropy and Results of Ablation Study}

\paragraph{Information Entropy}
\begin{table*}[h]
    \centering
    \begin{tabular}{ccc}
        \hline
        \textbf{Remove} & \textbf{Graph Entropy} & \textbf{Ratio (\%)} \\
        \hline
        Full  & 7.48 & 89.95 \\
        \hline
        clinical relationship  & 7.42  & 89.07 \\
        temporal events & 7.33  & 88.07 \\
        posology relationship  & 7.15 & 85.80 \\
        bodypart-directions  & 7.47  & 89.68 \\
        bodypart-problem  & 7.48  & 89.80 \\
        \hline        
        problem  & 6.80  & 81.62 \\
        treatment  & 7.27  & 87.25 \\
        test  & 7.27  & 87.30 \\
        drug  & 7.36  & 88.40 \\
        \hline
    \end{tabular}
    \caption{Results of the ablation study on information entropy: impact of removing each type of relationship or entity (MIMIC-III Full).}
    \label{tab:ablation}
\end{table*}
Information entropy, introduced by Shannon in 1948, is a fundamental concept in information theory that measures information loss by quantifying the difference between the expected information and the reduced information.
The entropy $H$ of a discrete source $X$ is given by:
\begin{equation}
H(X) = - \sum_{x \in X} P(x) \log_2 P(x).
\end{equation}
The entropy of text and serialised graph are calculated as follows:
\begin{equation}
H_\text{text} = - \sum_{x \in X_\text{text}} P_\text{text}(x) \log_2 P_\text{text}(x),
\end{equation}
\begin{equation}
H_\text{graph} = - \sum_{x \in X_\text{graph}} P_\text{graph}(x) \log_2 P_\text{graph}(x).
\end{equation}
The ratio of information loss \( L \) is defined as:
\begin{equation}
L = \frac{H_\text{text} - H_\text{graph}}{H_\text{text}}.
\end{equation}
\paragraph{Ablation Study}
Table \ref{tab:ablation} displays the information entropy results for different graphs after removing one type of relationship or entity. The \emph{`Text Entropy'} is 8.33 across all experiments. Notably, the removing \emph{`posology relationship'} and \emph{`problem'} have the most significant impact on the results. This analysis emphasizes the loss of textual information, whereas the ablation study in the main content examines the impact on ICD coding.
\subsection{Implementation Details and Results of Various DGCNN Configurations}

Table \ref{tab:parameters} outlines the hyperparameter settings for both the PLM-ICD baseline and our model. Our model requires a batch size of 1 per process, as we do not adjust the graph representation using padding, unlike typical text inputs. Due to computational resource constraints, we do not use the optimal hyperparameters for PLM-ICD. However, we maintain consistent hyperparameters within their shared architecture to ensure a fair comparison. The value of DGCNN indicates the size of the node representation for each convolution layer. A single DGCNN layer with a size of 768 achieves the best performance on the full dataset, while two DGCNN layers, each with a size of 384, performs best on the Top-50 dataset. Additionally, we initialize the node representation in the first layer using RoBERTa-base.

Tables \ref{tab:config1} and \ref{tab:config2} present additional experimental results for different configurations of the DGCNN architecture. The experiments utilize a complete graph with five types of relationships. In our initial experiment, we fix the final node size at 768 and compare the performance of DGCNN with different numbers of layers. The results indicate minimal performance differences between multi-layer and single-layer DGCNN models. However, models with evenly distributed layer sizes show slightly better performance. We also conduct experiments by varying the final node size and incrementally adding layers, each with an embedding size of 384. The results reveal an initial increase in performance, which subsequently decreases, with the optimal performance observed using two layers. Additionally, a similar trend is evident in a third experiment, which investigates varying sizes for each layer.
\subsection{Patient visualisation - Code 38.91}
\begin{figure}[t]
    \centering
    \includegraphics[width=\columnwidth]{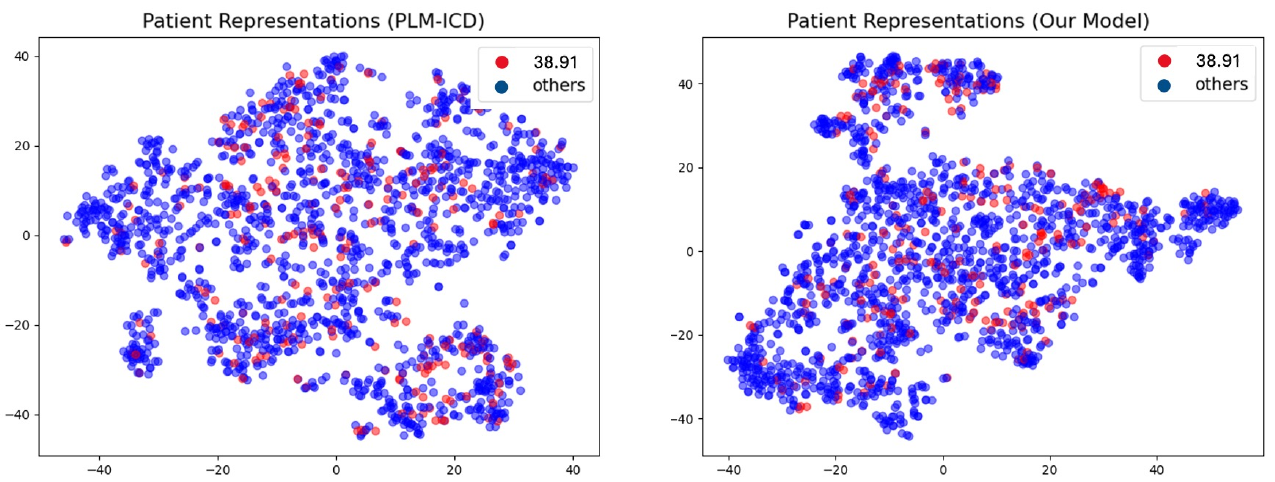}
    \caption{Visualisation of label-specific patients representation of code \texttt{38.91} \emph{`Arterial Catheterization'}, without (left) and with (right) using knowledge graphs as input. Instances with the corresponding ground-truth label are \textcolor{red}{red}.}
    \label{fig:bad_example}
\end{figure}

\begin{figure*}
    \centering
    \includegraphics[width=1.3\textwidth, angle=90]{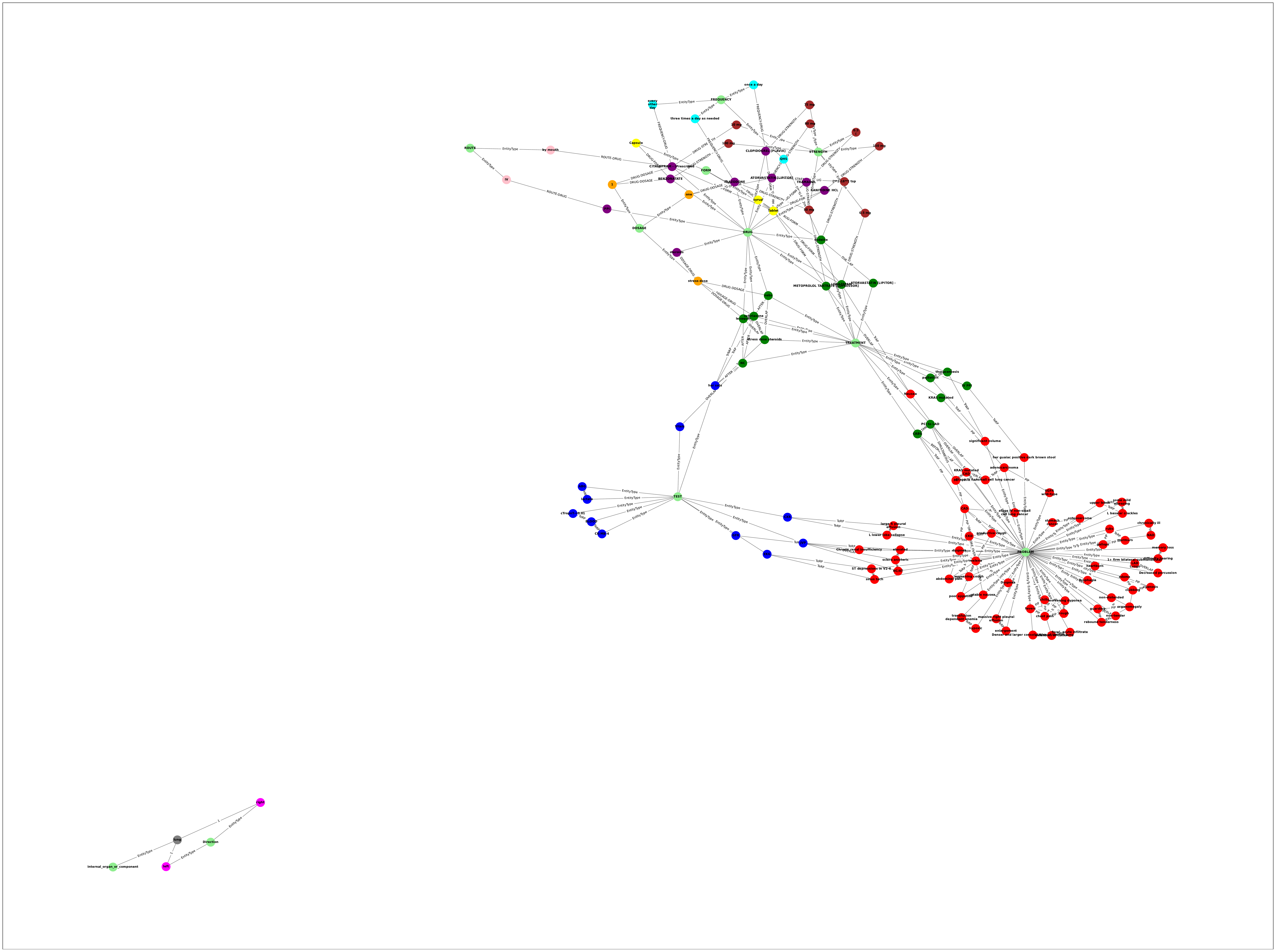}
    \caption{Visualisation of a Patient-Level Knowledge Graph.}
    \label{fig:visualisation}
\end{figure*}

We present a negative example of patient visualization for code \texttt{38.91} \emph{`Arterial Catheterization'} in Figure \ref{fig:bad_example}, where both models exhibit poor performance. Our model achieves an F1-score of 38.14\%, compared to 18.89\% for PLM-ICD.
\begin{table*}[ht]
    \centering
    \begin{tabular}{cll}
        \hline
        \textbf{RE Model} & \multicolumn{1}{c}{\textbf{Entity}} & \multicolumn{1}{c}{\textbf{Relationship}} \\
        \hline
        clinical relationship & PROBLEM, TREATMENT, TEST & TrAP: TREATMENT-PROBLEM\\
& & TeRP: TEST-PROBLEM\\
& & TrIP: TREATMENT-PROBLEM\\
& & TrCP: TREATMENT-PROBLEM\\
& & TeCP: TEST-PROBLEM\\
& & TrWP: TREATMENT-PROBLEM\\
& & PIP: PROBLEM-PROBLEM\\
& & O: No Relationship\\

        \hline
        temporal events & EVIDENTIAL, OCCURRENCE , DATE, & BEFORE, AFTER, OVERLAP \\
        & TREATMENT, TIME, ADMISSION,  & \\
        & TEST, FREQUENCY, CLINICAL\_DEPT,  & \\
        & DURATION, PROBLEM, DISCHARGE & \\
\hline
posology relationship & drug, dosage, duration, strength, frequency & DOSAGE-DRUG\\
& & DRUG-DURATION\\
& & DRUG-STRENGTH\\
&  & DRUG-FREQUENCY\\
\hline
bodypart-directions & direction-external\_body\_part\_or\_region,&1,0\\
&external\_body\_part\_or\_region-direction,&\\
&direction-internal\_organ\_or\_component,&\\
&internal\_organ\_or\_component-direction&\\
\hline
bodypart-problem & link between external\_body\_part\_or\_region\ & 1,0\\
 & or internal\_organ\_or\_component & \\
  &  and diseases entities (cerebrovascular\_disease & \\
 & , communicable\_disease, diabetes...) & \\
    \hline
    \end{tabular}
    \caption{Entities and relationships that RE models can extract.}
    \label{tab:RE_models}
\end{table*}

\begin{table*}
    \centering
    \begin{tabular}{ccc}
    \hline
    \textbf{Input} & \textbf{Parameter}&   \textbf{Value}\\
    \hline
     & number of processes & 4\\
         &train/evaluation batch size&  1\\
   Common  &gradient accumulation steps& 1\\
     &train epochs & 20 (Full) / 10 (Top-50)\\
     &warmup steps& 2000\\
     &random seed & 42\\
    \hline
     &max length&  5120\\
    Text &chunk size& 512\\
     &model mode & LAAT\\
     &pretrained model (text) & RoBERTa-base-PM\\
     \hline
      & DGCNN & 768 (Full) / 384-384 (Top-50)\\
    Graph & pretrained model (node) & RoBERTa-base\\
    \hline
    \end{tabular}
    \caption{Parameter settings for PLM-ICD (Common + Text) and our model (Common + Text + Graph) on the MIMIC-III Full and Top-50 datasets.}
    \label{tab:parameters}
\end{table*}

\begin{table*}
    \centering
    \begin{tabular}{clccccccc}
    \hline
    & & \multicolumn{2}{c}{\textbf{F1}} & & \multicolumn{2}{c}{\textbf{AUC}} & \textbf{Precision} & \textbf{Recall} \\ \cmidrule{3-4} \cmidrule{6-7}
     \textbf{Model} & \textbf{Embedding Size} & Macro & Micro & & Macro & Micro & P@8 & R@8 \\
         \hline
        1 layer & 768 & 11.05 & 59.72 & & 92.37 & 98.75 & 76.59 & 40.52 \\
        \hline
         & 256-512 & 10.69 & 59.52 & & 92.42 & 98.78 & 76.79 & 40.56 \\
        2 layers & 384-384 & 10.98 & 59.64 & & 92.65 & 98.83 & 76.63 & 40.53 \\
        \hline
         & 128-256-384 & 10.42 & 59.51 & & 92.55 & 98.79 & 76.47 & 40.39 \\
        3 layers & 256-256-256 & 10.53 & 59.70 & & 92.47 & 98.84 & 76.81 & 40.56 \\
        \hline
         & 128-128-256-256 & 10.46 & 59.47 & & 92.23 & 98.78 & 76.58 & 40.37 \\
         4 layers & 192-192-192-192 & 10.58 & 59.21 & & 92.14 & 98.78 & 76.47 & 40.37 \\
         \hline
        \hline
        1 layer & 384 & 10.77 & 59.77 & & 92.30 & 98.77 & 76.88 & 40.62 \\
        2 layers & 384-384 & 10.82 & 59.43 & & 92.54 & 98.78 & 76.63 & 40.53 \\
        3 layers & 384-384-384 & 10.49 & 59.37 & & 92.35 & 98.75 & 76.11 & 40.15 \\
        4 layers & 384-384-384-384 & 10.46 & 59.58 & & 92.23 & 98.74 & 76.79 & 40.55 \\
        \hline
        \hline
        1 layer & 128 & 10.23 & 59.06 & & 92.22 & 98.82 & 76.65 & 40.46 \\
        2 layers & 128-256 & 10.60 & 59.69 & & 92.47 & 98.83 & 76.85 & 40.57 \\
        3 layers & 128-256-384 & 10.42 & 59.51 & & 92.55 & 98.79 & 76.47 & 40.39 \\
        4 layers & 128-256-384-512 & 10.47 & 59.31 & & 92.21 & 98.76 & 76.30 & 40.27 \\
        \hline
    \end{tabular}
    \caption{Performance of Various DGCNN Architecture Configurations (MIMIC-III Full).}
    \label{tab:config1}
\end{table*}

\begin{table*}
    \centering
    \begin{tabular}{clccccccc}
    \hline
    & & \multicolumn{2}{c}{\textbf{F1}} & & \multicolumn{2}{c}{\textbf{AUC}} & \textbf{Precision} & \textbf{Recall} \\ \cmidrule{3-4} \cmidrule{6-7}
     \textbf{Model} & \textbf{Embedding Size} & Macro & Micro & & Macro & Micro & P@5 & R@5 \\
         \hline
        1 layer & 768 & 66.64 & 71.37 & & 91.77 & 94.16 & 66.52 & 64.33 \\
        \hline
         & 256-512 & 67.64 & 71.72 & & 92.12 & 94.30 & 66.82 & 64.74 \\
        2 layers & 384-384 & 67.81 & 71.63 & & 92.04 & 94.22 & 67.08 & 65.11 \\
        \hline
         & 128-256-384 & 66.30 & 70.94 & & 91.71 & 93.98 & 66.58 & 64.39 \\
        3 layers & 256-256-256 & 67.54 & 71.83 & & 92.19 & 94.37 & 67.04 & 65.14 \\
        \hline
         & 128-128-256-256 & 66.79 & 71.39 & & 92.28 & 94.31 & 66.87 & 64.92 \\
         4 layers & 192-192-192-192 & 67.67 & 72.03 & & 92.30 & 94.44 & 66.79 & 65.07 \\
         \hline
                 \hline
        1 layer & 384 & 66.91 & 71.12 & & 92.04 & 94.19 & 66.47 & 64.63 \\
        2 layers & 384-384 & 67.81 & 71.63 & & 92.04 & 94.22 & 67.08 & 65.11 \\
        3 layers & 384-384-384 & 66.89 & 71.41 & & 92.26 & 94.32 & 67.09 & 65.20 \\
        4 layers & 384-384-384-384 & 66.55 & 71.24 & & 92.15 & 94.37 & 66.86 & 64.85 \\
        \hline
                \hline
        1 layer & 128 & 66.62 & 70.79 & & 91.89 & 94.10 & 66.50 & 64.45 \\
        2 layers & 128-256 & 67.63 & 71.72 & & 92.00 & 94.25 & 66.71 & 64.63 \\
        3 layers & 128-256-384 & 66.30 & 70.94 & & 91.71 & 93.98 & 66.58 & 64.39 \\
        4 layers & 128-256-384-512 & 65.80 & 71.34 & & 92.05 & 94.32 & 66.14 & 64.22 \\
        \hline
    \end{tabular}
    \caption{Performance of Various DGCNN Architecture Configurations (MIMIC-III Top-50).}
    \label{tab:config2}
\end{table*}
\end{document}